\pdfoutput=1

\documentclass[11pt]{article}

\usepackage[final]{acl}

\usepackage{times}
\usepackage{latexsym}
\usepackage{amsmath}
\usepackage{amsfonts}
\usepackage{multirow}
\usepackage{enumitem}
\usepackage{array}
\usepackage{soul,color,xcolor}
\usepackage{hyperref}
\definecolor{revisioncolor}{rgb}{0.8, 0, 0}
\newcommand{\revision}[1]{\textcolor{revisioncolor}{#1}}
\renewcommand{\revision}[1]{#1}
\newcommand{\revisionapril}[1]{\textcolor{revisioncolor}{#1}}
\renewcommand{\revisionapril}[1]{#1}

\usepackage[T1]{fontenc}

\usepackage[utf8]{inputenc}

\usepackage{microtype}

\usepackage{inconsolata}

\usepackage{graphicx}

\title{GiLT: Augmenting Transformer Language Models with Dependency Graphs}

\author{Tianyu Huang, Yida Zhao, Chuyan Zhou, Kewei Tu\thanks{\; Corresponding Author}\\
  School of Information Science and Technology, ShanghaiTech University \\
  Shanghai Engineering Research Center of Intelligent Vision and Imaging\\ 
    {\tt \{huangty2024,zhaoyd2023,zhouchy2022,tukw\}@shanghaitech.edu.cn}\\
 }

\begin{document}
\maketitle
\begin{abstract}
Augmenting Transformers with linguistic structures effectively enhances the syntactic generalization performance of language models. Previous work in this direction focuses on syntactic tree structures of languages, in particular constituency tree structures. We propose \emph{Graph-Infused Layers Transformer Language Model} (GiLT) which leverages dependency graphs for augmenting Transformer language models. Unlike most previous work, GiLT does not insert extra structural tokens in language modeling; instead, it injects structural information into language modeling by modulating attention weights in the Transformer with features extracted from the dependency graph that is incrementally constructed along with token prediction. 
In our experiments, GiLT with semantic dependency graphs achieves better syntactic generalization while maintaining competitive perplexity in comparison with Transformer language model baselines.
In addition, GiLT can be finetuned from a pretrained language model to achieve improved downstream task performance.
\revisionapril{Our code is released at \href{https://github.com/cookie-pie-oops/GiLT-LM}{https://github.com/cookie-pie-oops/GiLT-LM}.}
\end{abstract}

\section{Introduction}
    Transformer language models (LMs) have shown excellent performance in language modeling and downstream tasks \citep{vaswaniAttentionAllYou2017}. Notably, linguistic structures such as syntactic and semantic parses that have been deemed essential in traditional natural language processing are absent from the model design and training process of Transformer LMs.

    Over the past decade, a number of researchers have been trying to integrate linguistic structures into neural language models. Among them are syntactic LMs which jointly model syntactic structures and surface words \citep{choe-charniak-2016-parsing}. These include earlier work such as RNNG, which combines constituency parsing with recurrent neural networks \citep{dyer-etal-2016-recurrent, kim-etal-2019-unsupervised, noji-oseki-2021-effective}, and recent studies that incorporate constituency and dependency syntax into Transformers \citep{yoshida-oseki-2022-composition, qian-etal-2021-structural, transformer-grammars, murty-etal-2023-pushdown, zhao-etal-2024-dependency}. \revision{Empirically, they achieve stronger syntactic generalization compared with standard Transformer LMs while retaining competitive language modeling performance. }

    However, existing researches in this direction have two major limitations. 
    First, most of them are based on constituency syntactic tree structures. Dependency tree structures, another important form of syntax, receive much less attention \citep{zhao-etal-2024-dependency}. In addition, little work has been done to jointly model linguistic structures other than syntactic trees in LMs.
    Second, most of the existing methods require inserting additional tree-building operations into the input and output sequence, leading to longer sequence lengths and higher computational cost, and making it harder to finetune a pretrained LM into a syntactic LM. An exception is Pushdown Layers \citep{murty-etal-2023-pushdown}, which leverages syntactic trees to guide attention computation without changing the LM's symbol space.

    In this paper, we propose the Graph-Infused Layer Transformer LM (GiLT) that addresses the above-mentioned limitations in integrating linguistic structures into Transformer LMs. GiLT is based on \emph{dependency graphs} that subsume both syntactic dependency trees and \emph{semantic} dependency graphs, thus extending syntactic LM research beyond syntax. Inspired by Pushdown Layers \citep{murty-etal-2023-pushdown}, GiLT incrementally constructs dependency graphs without changing the symbol space of the underlying LMs, and modulates attention scores with features computed from graph attributes such as node degrees, depths and distances.
    
    \revision{Experimental results show that GiLT achieves gains in syntactic generalization over baselines with almost no degradation in perplexity on language modeling.} Furthermore, GiLT finetuned from a pretrained GPT2 achieves better performance on downstream tasks compared with the original pretrained GPT2, suggesting that the Graph-Infused layer is a competitive alternative to standard self-attention.

 \revision{In summary, our contributions are as follows:
 \begin{itemize}[itemsep=0pt, topsep=3pt]
     \item We propose \emph{Graph-Infused Layers}, leveraging dependency graphs to enhance LMs by our novel \emph{graph-based feature tapes} without modifying the input or output space.
     \item 
     Comprehensive experiments on language modeling, syntactic evaluation and finetuning on text classification demonstrate competitive perplexity, improved syntactic generalization and language understanding.
     Ablation study on feature tapes shows the importance of each part and test on generation speed illustrates the advantage of not requiring extra tokens.
 \end{itemize}}

\section{Background}
\subsection{Pushdown Layers}
Transformer LM with Pushdown Layers \citep[Pushdown-LM,][]{murty-etal-2023-pushdown} is a type of syntactic LMs that incrementally builds constituency syntactic trees and modulates attention scores based on the constituency trees. Unlike other syntactic LMs, it does not change the symbol space of the underlying LM.

At each decoding step $i$, Pushdown-LM predicts shift/reduce operations to simulate the status of a pushdown automaton that corresponds to the partially built constituency tree, and records on a \emph{stack tape} $\mathbf{t}_i$ the depths of all the tokens that are already generated in the partially built constituency tree.

Pushdown-LM then augments self-attention with stack tape $\mathbf{t}_i$:
\begin{equation}
    \label{eq:attention_pushdown}
    \tilde{\alpha}^l_{ij} = [\mathbf{h}^l_j + \mathbf{d}_{ij}^l]^\top \mathbf{W}_{k}^{\top} \mathbf{W}_{q}\mathbf{h}_i^l
\end{equation}
where $\tilde{\alpha}_{ij}^l$ is the attention score before softmax assigned to the $j$-th token from the $i$-th token at layer $l$, $\mathbf{h}^l_j$ is the hidden state of $j$-th token at the $l$-th attention block, $\mathbf{d}_{ij}^l$ is the embedding of the depth of the $j$-th token recorded in $\mathbf{t}_i$, and $\mathbf{W}_{k}$ and $\mathbf{W}_{q}$ are learnable query and key matrices in self-attention.
In this way, structural information from the constituency tree is implicitly introduced into self-attention computation and thus influences the decoding of the underlying LM.

\subsection{Semantic Dependency Graphs}
A semantic dependency graph forms as a directed acyclic graph instead of a tree. The dependencies in the graph, where nodes correspond to words, illustrate semantic relations (\emph{e.g.}, agent and patient  in \citealt{10.1162/0891201053630264}). The graph often includes a virtual root node.

In this paper, we consider three types of semantic dependency graphs from \citet{oepen-etal-2015-semeval} as discussed below.
DELPH-IN MRS-Derived Bi-Lexical Dependencies \citep[DM,][]{pub6619} are derived from Deep Bank \cite{DBLP:journals/nle/Flickinger00}, in which roots designate the highest-scoping predicate in the graph. Enju Predicate–Argument Structures (PAS) originate from Enju Treebank \cite{2006FromLT}, which is obtained by automatically annotating the PTB. The root of PAS denotes the semantic head in the sentence. Prague Semantic Dependencies (PSD) are based on Prague Czech-English Dependency Treebank \cite{Hajic2012AnnouncingPC}, where the roots mostly correspond to main verbs.

\section{Graph-Infused Layers}
\revision{We introduce a dependency-graph-based language model, \emph{Graph-Infused Layers Transformer LM} (GiLT), which simultaneously generates tokens that form sentences, and dependencies that incrementally construct dependency graphs over the sentences.
We first score possible dependencies that link the current word to previous words (Section \ref{dep-scoring}), then update the dependency graph based on the scoring (Section \ref{graph update}), and utilize the \emph{graph-based feature tapes} (Section \ref{Node Relative Position}), which characterize generated tokens in the graph, to modulate attention computation (Section \ref{GiLT-score}).}

\subsection{Dependency Scoring}
\label{dep-scoring}

\revision{Whenever a word $w_i$ is generated by the Transformer LM, we score all possible dependencies connected from and to $w_i$ with a biaffine mechanism. Since a word may correspond to multiple tokens, we first define the word-level representations that serve as input to the biaffine module.}

\revision{Suppose a word $w_i$ is tokenized into $m$ tokens with input embeddings $\{\mathbf{x}_{k},\cdots, \mathbf{x}_{k+m-1}\}$ and corresponding hidden states $\{\mathbf{h}_{k}^l,\cdots, \mathbf{h}_{k+m-1}^l\} \subseteq \mathbb{R}^d$ from all layers $l=1,\dots,L$. We define its word-level representation $\mathbf{o}_i\in\mathbb{R}^{3d}$ by concatenating three components: (i) the hidden state from the middle layer,   $\mathbf{h}_{k-1}^{L/2}$; (ii) the hidden state from the penultimate layer, $\mathbf{h}_{k-1}^{L-1}$; (iii) the input embedding of the first token, $\mathbf{x}_k$, which provides direct lexical information about the word. According to the assumption in \citet{murty-etal-2023-pushdown}, since $\mathbf{h}_{k-1}^{L/2}$ and $\mathbf{h}_{k-1}^{L-1}$ are hidden states from sufficiently deep layers used to predict the $k$-th token, they capture useful information about the $k$-th token rather than the ($k-1$)-th token. We do not use the final-layer hidden states to reserve them exclusively focused on next token prediction.}
Note that we do not use input embeddings and hidden states computed after $\mathbf{x}_k$ so that we can predict all dependencies of $w_i$ before processing $\mathbf{x}_k$, thus being able to infuse structural information from the dependency graph to the hidden states of the tokens in $w_i$. 

\begin{figure}[t]
    \centering
    \includegraphics[width=0.8\columnwidth]{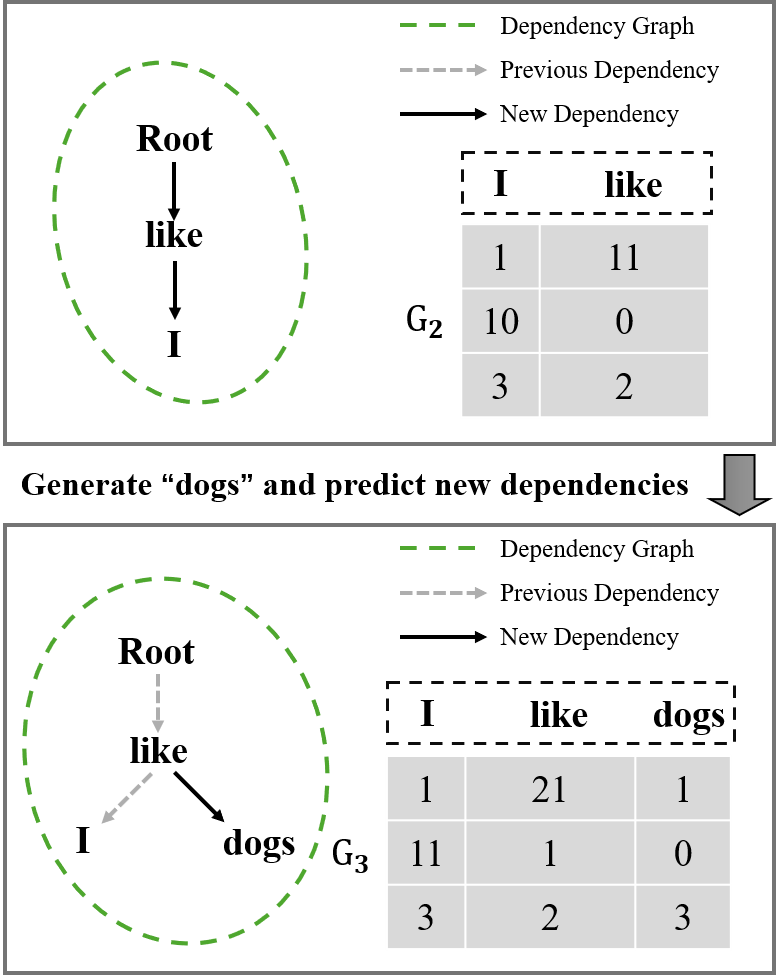}
        \caption{Illustration of how the feature tape is recomputed when generating a sentence and constructing its dependency graph. Rows in $G_2$ and $G_3$ from top to bottom correspond to Degree, Distance and Depth, respectively. We set $m_{in}=1$ and $m_{out}=10$ for this example. As \emph{dogs} is predicted, one dependency is added to the graph.}
    \label{fig:pred depth}
\end{figure}

We follow the biaffine parsing approach \citep{dozat-manning-2018-simpler}
to compute the probability $p_{ij}$ of the dependency from the word $w_i$ to $w_j$. 
Note that for the root node, we use a learnable vector as its word representation:
\begin{equation}
  \label{eq:biaffine}
  \begin{aligned}
    \tilde{\mathbf{o}}^{par}_i &= \text{MLP}_{par}^2(\text{MLP}_{par}^1(\mathbf{o}_i) + \mathbf{pe}_{ii}) \\
    \tilde{\mathbf{o}}^{chd}_j &= \text{MLP}_{chd}^2(\text{MLP}_{chd}^1(\mathbf{o}_j) + \mathbf{pe}_{ij}) \\
    p_{ij} &= \sigma(
    \tilde{\mathbf{o}}^{par}_{i}{}^{\top}
    \mathbf{W}_p{\mathbf{\tilde{o}}^{chd}_j})
  \end{aligned}
\end{equation} 
where $\mathbf{o}_i$ is the word representation of $w_i$ as defined above, $\mathbf{W}_{par} \in \mathbb{R}^{d\times d}$ is a learnable matrix, $\text{MLP}^{1/2}_{{par}/{chd}}$ denotes the first/second MLP for computing parent/child representations $\tilde{\mathbf{o}}_i^{{par}/{chd}} \in \mathbb{R}^{d}$, $\sigma$ denotes the sigmoid function, and $\mathbf{pe}_{ij}$ denotes the positional embedding which is a sum of the sinusoid encoding of $|i-j|$ and the embedding of graph-based feature tape $G_i$ (see Section~\ref{Node Relative Position}). 

\subsection{Graph Update}
\label{graph update}

Given dependency probabilities $\{p_{ij},p_{ji},p_{ii}\}$ where $j\in \{0,\cdots,i-1\}$ for all possible dependencies with regard to the $i$-th word $w_i$, a straightforward method to greedily update the dependency graph is to add any dependency whose probability exceeds 0.5. 
However, this becomes computationally intractable when we employ beam search of dependency graphs (Section~\ref{inference}) 
because of the exponentially large search space. To address this issue, we consider a restricted subspace of dependency graphs by using a two-step method as follows. For $w_i$:

(i) \revision{We predict the number of dependencies $c_i \in \{0, 1, \cdots, C\}$, where $C$ is a constant upper bound:}
\begin{equation}
    \label{action}
    \begin{aligned}
        \mathbf{s} &= \sum_{j = 0}^{i} \tilde{\mathbf{o}}^{par}_i \odot \mathbf{W}_s \tilde{\mathbf{o}}^{chd}_j + \sum_{j = 0}^{i-1} \tilde{\mathbf{o}}^{par}_j \odot \mathbf{W}_s \tilde{\mathbf{o}}^{chd}_i\\
        \boldsymbol{\pi}_i &:= [\pi_i^0, \pi_i^1, 
        \cdots, \pi_i^C]^T \\
        &= \text{softmax}\left(\mathbf{W}_{a}^\top(\frac{\mathbf{s}}{\sqrt{2i - 1}}) + \mathbf{b}_{a}\right) \\
    \end{aligned}
\end{equation} 
where $\odot$ denotes the Hadamard product operation, $\mathbf{W}_a \in \mathbb{R}^{(C+1) \times d}$, $\mathbf{W}_s \in \mathbb{R}^{d\times d}$ and $\mathbf{b}_a \in \mathbb{R}^{C+1}$ are learnable parameters, $\boldsymbol{\pi}_i \in \mathbb{R}^{C+1}$ is the probability distribution over $\{0, 1, \cdots, C\}$. To normalize the variance, we scale $\mathbf{s}$ by $\sqrt{2i - 1}$.

(ii) 
\revisionapril{The value of $c_i$ can be obatined through either greedy decoding (choosing the most probable value) or sampling from $\boldsymbol{\pi}_i$}. Then we pick $c_i$ highest-scoring dependencies and add them to the dependency graph. This two-step method reduces the search space from exponential to linear for each step in beam search.

\subsection{Feature Extraction}
\label{Node Relative Position}
\revision{Given the input sequence $x_{<k}$, where $x_k$ is the first token of the $i$-th word $w_i$ as defined in Section~\ref{dep-scoring}, we extract features from the partially constructed dependency graph and form a graph-based feature tape $G_k=[g_{1k}, g_{2k}, \cdots, g_{kk}] \in \mathbb{N}^{3 \times k}$ for the token $x_k$. Note that the graph is word-level, but $G_k$ corresponds to a token. Therefore, for any token $i,j$ belongs to the same word, $g_{ik} = g_{jk}$ in $G_k$.}


The feature tape involves three graph-based features: (i) degree, an attribute for each word; (ii) distance, measuring connectivity between words; (iii) depth, reflecting the global structure of the graph.

\paragraph{Degree.} The degree of a word refers to the number of its incoming and outgoing dependencies, denoted as $c_{in}$ and $c_{out}$ respectively. 
According to the definition, $c_{in} + c_{out}$ is the degree for each word, but empirically, we discover that weighted summation achieves better performance (see section \ref{sec:ablation}): we assign weight $m_{in} \in \mathbb{Z}^+$ to in-degree and $m_{out} \in \mathbb{Z}^+$ to out-degree, where $0<m_{in} < m_{out}$, and set the degree as $m_{out} c_{out} + m_{in} c_{in}$.

\paragraph{Distance.} 

\revision{The distance from word $w_i$ to word $w_j$ is computed by finding the weighted shortest path on the current dependency graph. When traversing a dependency \revisionapril{along} its direction, we weight it by $m_{out}$; against the direction, we use $m_{in}$. Thereby encoding dependency direction information into the distance measure.}
Intuitively, the distance measures the relevance of the two words. Specifically, distance recorded in $g_{1k}$ is measured from the word of token $x_k$ to the word of token $x_1$.

\paragraph{Depth.} 
\revision{The dependency graphs used in our work are all rooted. We define the depth of a word to be the length of the shortest path to the root plus one in the undirected backbone of the dependency graph. Since the dependency graph is partial, a word may be disconnected from the root and we set its depth to be 0. We can compute the depths of all the words with breadth-first search starting from the root. A visited flag ensures that each word is processed exactly once.}

\revision{In Figure \ref{fig:pred depth}, we illustrate the feature tapes when generating an example sentence.}

\subsection{Computing Attention Scores}
\label{GiLT-score}
We incorporate information in the graph-based feature tape $G_k$ into the Transformer LM by modifying the self-attention module. Specifically, we first map $G_k \in \mathbb{N}^{3 \times k}$ via a learned embedding layer onto a global embedding $\tilde{\mathbf{e}}_{k} \in \mathbb{R}^{3 \times k \times \tilde{d}}$. For each Transformer layer $l$, we apply a linear projection $f_l$ for feature fusion, that is, $\mathbf{e}^{l}_{k} = f_l(\tilde{\mathbf{e}}_{k}) \in  \mathbb{R}^{k \times d}$. For each token position $j \in \{0, 1, \cdots, k\}$, the corresponding fused graph feature $\mathbf{e}^{l}_{kj} \in \mathbb{R}^{d}$ is directly added to its key in attention computation,
\begin{equation}
    \label{eq:attention_GiLT}
    \tilde{\alpha}^l_{kj} = [\mathbf{h}^l_j + \mathbf{e}^{l}_{kj}]^\top \mathbf{W}_{k}^{\top} \mathbf{W}_{q}\mathbf{h}_k^l
\end{equation}
In this work, we follow the practice of Transformer-XL \citep[TXL,][]{dai-etal-2019-transformer} for attention computation, so we additionally transform Equation~\ref{eq:attention_GiLT} as follows. 
\begin{equation}
    \label{txl-GiLT}
    \begin{aligned}
        {\tilde{\alpha}}^l_{kj} &= {\mathbf{h}^l_{j}}^\top \mathbf{W}_{k,c}^\top \mathbf{W}_q \mathbf{h}^l_{k} \\
        &+ (\mathbf{r}_{kj}+\mathbf{e}^{l}_{kj})^\top \mathbf{W}_{k,r}^\top \mathbf{W}_q {\mathbf{h}^l_{j}} \\
        &+ u^\top \mathbf{W}_{k,c} \mathbf{h}^l_{k} + v^\top \mathbf{W}_{k,r} (\mathbf{r}_{kj}+\mathbf{e}^{l}_{kj}) \text{,}
    \end{aligned}
\end{equation}
where $\mathbf{r}_{kj}$ is a vector with sinusoid encoding of $|k-j|$, $W_{k,c}$ and $W_{k,r}$ are key matrix for respectively extracting content and relative representation, $u$ and $v$ are learnable bias vectors. 

\subsection{Training and Inference}
\paragraph{Training.} Given a corpus of strings annotated with dependency graphs, we precompute graph-based feature tape $G_k$ for each prefix $x_{\leq k}$ based on the ground-truth dependencies over $x_{\leq k}$. After this preprocessing step, GiLT can be trained in parallel like a standard Transformer. During training, teacher forcing is applied to both token prediction and dependency prediction. Given a string $x$ with $N$ tokens and $M$ words and its ground-truth dependency graph, 
\revision{we derive a sequence of one-hot ground-truth vectors of length $M$ $\{\hat{\boldsymbol{\pi}}_1, \cdots, \hat{\boldsymbol{\pi}}_M\}$ indicating the number of dependencies of each word}
and a matrix $\hat{p}$ where $\hat{p}_{ij} = 1$ iff. there is a dependency from word $i$ to word $j$.
The training loss function is defined as follows:
\begin{equation}
    \begin{aligned}
    \mathcal{L} &= \alpha \frac{1}{N} \sum_{i=1}^N \text{CE}(x_i, \hat{x}_i) \\ &+ \beta \frac{1}{M^2} \sum_{j=1}^{M} \sum_{k=1}^{M} \text{BCE}(p_{jk}, \hat{p}_{jk})\\
    &+ \gamma \frac{1}{M} \sum_{k=1}^{M} \text{CE}(\boldsymbol{\pi}_k, \hat{\boldsymbol{\pi}}_k)
    \end{aligned}
\end{equation}
where $\alpha$, $\beta$, $\gamma$ are constant coefficients, $p_{jk}$ and $\boldsymbol{\pi}_k$ are from Eq. ~\ref{eq:biaffine} \& \ref{action} respectively.

\paragraph{Inference.}\label{inference} 
GiLT jointly generates a string $x$ and its dependency graph $y$. As discussed in Section \ref{graph update}, GiLT considers only a subspace $Y$ of dependency graphs and models uncertainty over the subspace with uncertainty over numbers of dependencies of the words in $x$. Therefore, the joint probability $p(x,y)$ is computed as follows.
\begin{equation}
    \label{prob}
    \begin{aligned}
        p(x,y) &= \prod_{k=1}^N p(x_k|x_{<k}; G_{k-1}(y))
         \\ & \times  \prod_{j=1}^M p(c_j(y)|x_{\leq z_j}; G_{z_j}(y))
    \end{aligned}
\end{equation}
where $N$ and $M$ are the numbers of  tokens and words in $x$, $c_j(y)$ is the number of dependencies of the $j$-th word in graph $y$, $z_j$ is the index of the last token within the $j$-th word, $G_{k}(y)$ is the feature tape for token $x_k$ based on the subgraph of $y$ containing only the words corresponding to the first $k$ tokens. 

Computing the marginal $p(x) = \sum_{y \in Y} p(x,y)$ is computationally intractable due to the huge space of all possible graphs. Following \citet{murty-etal-2023-pushdown}, we approximate it by marginalizing (summing) over a relatively small set of dependency graphs produced via beam search (i.e., retaining multiple most likely values of $c_j$ and their corresponding dependencies for every beam). The approximated marginal probability in this way is an exact lower bound of its true value. Note that since GiLT does not generate extra tokens representing parsing actions in the output sequence, we do not need to use complicated word-synchronous beam search decoding \cite{stern-etal-2017-effective}, which has been widely used in previous syntactic LMs.

\section{Experiments}
\subsection{Sentence-Level Language Modeling}
\label{exp:lm}

\paragraph{Dataset and preprocessing.} We use the BLLIP-{\small LG} dataset of \citet{charniak_eugene_bllip_2000}, with training splits from \citet{hu-etal-2020-systematic}. We obtain annotated PSD, PAS and DM dependency graphs with unlabeled dependencies by parsing the dataset with ACE \cite{wang-etal-2021-automated}. Since a dependency tree can be seen as a special case of a dependency graph, we also obtain unlabeled projective dependency trees with the Biaffine-{\small RoBERTa} parser \cite{dozat-etal-2017-deep} following \citet{zhao-etal-2024-dependency}. Tokenization is performed with the same scheme as in \citet{transformer-grammars} with SentencePiece \cite{kudo-richardson-2018-sentencepiece}. We follow \citet{murty-etal-2023-pushdown} and model each sentence independently.

\paragraph{Setup.} We evaluate the perplexity (PPL) of the models on the BLLIP-{\small LG} dataset. We train a 16-layer TXL \cite{dai-etal-2019-transformer} language model of 252M parameters as the baseline. We also reimplement Pushdown-LM \cite{murty-etal-2023-pushdown} based on the code base of TXL for fair comparison. We use PSD, DM and PAS dependency graphs to train our GiLT respectively, resulting in three models: GiLT-PSD, GiLT-DM and GiLT-PAS. We also train GiLT on dependency parse trees, resulting in the GiLT-DP. \revisionapril{Meanwhile, we compare our models with constituency-based and dependency-based syntactic Transformer language models including: (i) Parsing as Language Model (PLM \& PLM-Mask) of \citet{qian-etal-2021-structural}, (ii) Transformer Grammars (TG) of \citet{transformer-grammars}, and (iii) Dependency Transformer Grammars (DTG) of \citet{zhao-etal-2024-dependency}.}

\begin{table}
  \centering
  \small
  \begin{tabular}{c|cccc}
    \hline
    \multicolumn{2}{c}{\textbf{Model}} & \textbf{PPL} $\downarrow$ & \textbf{10\%BLiMP} $\uparrow$ & \textbf{SG} $\uparrow$ \\
    \hline
    \multicolumn{5}{c}{\textbf{Models that add structural tokens to inputs}}\\
    \hline
    \multicolumn{2}{c}{PLM} & 29.8$^\spadesuit$ & 75.1$^\clubsuit$ & 76.9 \\
    \multicolumn{2}{c}{PLM-Mask} & 49.1$^\spadesuit$ & 75.3$^\clubsuit$ & 74.7 \\
    \multicolumn{2}{c}{TG} & 18.4 & 73.5$^\clubsuit$ & 79.6 \\
    \multicolumn{2}{c}{DTG} & 14.9 & \underline{76.1}$^\clubsuit$ & \underline{81.2} \\
    \hline
    \multicolumn{5}{c}{\textbf{Models that do not add extra tokens to inputs}}\\
    \hline
    \multicolumn{2}{c}{TXL (baseline)} 
    & 14.8 & 75.1 & 72.1 \\
    \multicolumn{2}{c}{TXL-Large (334M)} 
    & 14.7 & 75.4 & 71.8 \\
    \multicolumn{2}{c}{Pushdown-LM} 
    & \underline{\textbf{14.6}} & 74.0 & 74.6 \\
    \multirow{4}{*}{Ours}
    & GiLT-DP & 15.5 & \underline{\textbf{76.1}} & 72.3 \\
    & GiLT-PAS & 15.0 & 73.0 & 75.7 \\
    & GiLT-DM & 14.9 & 74.9 & 76.3 \\
    & GiLT-PSD & 14.9 & 75.7 & \textbf{79.7} \\
    \hline
  \end{tabular}
  \caption{Results on language modeling and syntactic generalization. The best values over models that do not add extra tokens are \textbf{bold}. Overall best values are \underline{underlined}. All PPL results except for that of TXL are approximated upper bounds. SG scores are computed without the "other" suite.
  $\spadesuit$ denotes that the PPL is taken from the original paper. $\clubsuit$ denotes that the result is evaluated with the full BLiMP dataset reported in the original paper.} 
  \label{tab:language modeling result}
\end{table}

We set the hyperparameters for GiLT with $m_{in}$ = $1$, $m_{out}$ = $10$, $\alpha = 1$, $\beta = 0.2$, $\gamma = 0.2$, $\tilde{d} = 256$, and $d = 1024$.
This results in 268M parameters for Transformer and 54M parameters for the modules described in Section~\ref{dep-scoring}-\ref{GiLT-score} in GiLT. For TXL, the same configuration of hyperparameters (model size, dropout, learning rate schedulers) is used as in \citet{zhao-etal-2024-dependency}. For perplexity computation, we apply beam search of dependency graphs with the same beam size $b$ of 300 following Pushdown-LM \cite{murty-etal-2023-pushdown} to estimate the perplexity upper bound of GiLT. \revision{To account for the additional parameters in GiLT in comparison with the baseline, we also train a TXL-Large with 22 layers (6 more than the base TXL, resulting in 334M parameters), to investigate the impact of scaling up model sizes alone.}

\paragraph{Result.} We report the PPL of all models in Table~\ref{tab:language modeling result}. 
\revision{Some syntactic language models, such as PLM and TG, introduce syntactic inductive bias at the cost of language modeling performance.
On the other hand, Pushdown-LM achieves the best PPL and even outperforms the baseline, confirming previous observation that Pushdown-LM excels in language modeling among syntactic LMs \cite{murty-etal-2023-pushdown}. 
The three GiLT models based on dependency graphs maintain PPL values comparable to both Pushdown-LM and the baseline.}
In contrast, GiLT-DP exhibits a higher PPL, highlighting the limitations of tree structures compared to more flexible graph-based structures.

\subsection{Syntactic Generalization}
We evaluate syntactic generalization on BLiMP \cite{warstadt-etal-2020-blimp-benchmark} and the SG test suites \cite{hu-etal-2020-systematic}.

\paragraph{Setup.} We use the same baseline models as in Section~\ref{exp:lm}. For BLiMP, models are evaluated on their ability to assign higher probability to grammatical sentences than to their minimally altered ungrammatical counterparts. The reported BLiMP score is the percentage of such pairs where the model succeeds, i.e., where the grammatical sentence receives a higher probability. Due to limited computational resources, we evaluate on a 10\% subset of the BLiMP dataset, selecting every tenth example (i.e., the 1st, 11th, 21st, etc.). We find that doing this only leads to very small perturbations to the scores (e.g., 0.2 for TXL).

\revision{The SG test suites include seven syntactic phenomenon classes. For each suite, the model is evaluated on its ability to satisfy a predefined inequality concerning the probability of generating a target span. We report the per-suite satisfaction percentage rate (i.e., the fraction of inequalities that hold) and then compute the SG score as the macro-average of six rates except for the “other” suite, as it contains only a single sentence, does not correspond to any specific syntactic phenomenon, and disproportionately influences the macro-average.}

For 10\%BLiMP and SG, we use beam search of dependency graphs to both compute marginal probability $p(x)$ and conditional probability $p(x_t|x_{<t})$.

\begin{figure}[t]
    \centering
    \includegraphics[width=\linewidth]{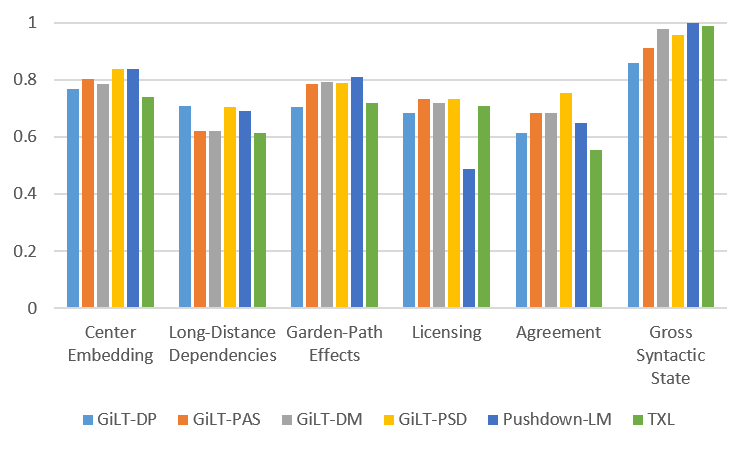}
    \caption{Scores on the 6 circuits of the SG test suites among models without extra tokens.}
    \label{fig:SG}
\end{figure}

\paragraph{Result.} 
\revision{The results are presented in Table~\ref{tab:language modeling result}. 
Although TXL-Large has more parameters than our models, its improvements are marginal, indicating that simply scaling up TXL without syntactic inductive bias fails to improve syntactic generalization.
GiLT-PSD outperforms most models in both tests, surpassing the baseline by 0.6 points in 10\%BLiMP and 7.6 points in SG. GiLT-DP achieves the best 10\%BLiMP performance, matching that of DTG.
In contrast, Pushdown-LM exhibits better SG performance but worse 10\%BLiMP performance than the baseline.}

\subsection{Finetuning on Pretrained LM}
Since GiLT does not change the symbol space of the Transformer LM, it can be finetuned from any pretrained language model on any datasets annotated with dependency graphs to introduce syntactic inductive bias. 
We therefore evaluate GiLT by starting from a pretrained GPT2 model and finetuning it on downstream tasks.
\revisionapril{Meanwhile, we are also curious about whether finetuned models still exhibit better syntactic generalization. Thus, we also evaluate them on BLiMP and the SG test suites.}

\begin{table}[b]
  \centering
  \small
  \setlength{\tabcolsep}{4pt}
  \begin{tabular}{ccccc}
    \hline
    \textbf{Model} & \textbf{RTE} $\uparrow$ & \textbf{SST2} $\uparrow$ & \textbf{MRPC} $\uparrow$ & \textbf{STS-B} $\uparrow$ \\
    \hline
    Post-GPT2 & 64.1 & 94.84 & 80.75/86.29 & 84.2/83.6 \\
    GiLT-GPT2 & \textbf{65.3} & \textbf{95.11} & \textbf{81.39/86.97} & \textbf{85.2/84.3} \\
    \hline
  \end{tabular}
  \caption{\revisionapril{Results when Post-GPT2 and GiLT-GPT2 are separately finetuned on each downstream task.}}
  \label{tab:finetune result}
\end{table}

\paragraph{Setup.} \revision{We use the pretrained GPT2-medium (355M) \cite{radford2019language} as the base model.
We create GiLT-GPT2 by replacing its last 12 vanilla transformer layers with our Graph-Infused layers and finetune GiLT-GPT2 on BLLIP-{\small LG}.
We evaluate the language understanding ability of GiLT-GPT2 on four downstream text classification tasks from GLUE \cite{wang-etal-2018-glue}: RTE, SST2, MRPC and STS-B. Each task is transformed into a language modeling task via prompting (details are provided in Appendix~\ref{sec:appendix finetune}).
We use GiLT-GPT2 to parse each prompt and then finetune the model with the parsed dependency graph on text classification.
For fair comparison, we also perform the same workflow for vanilla GPT2-medium, i.e., finetuning on BLLIP-{\small LG} \revisionapril{to obtain Post-GPT2,} and then further finetuning Post-GPT2 on downstream task data.}

\begin{table}[b]
  \centering
  \setlength{\tabcolsep}{4pt}
  \begin{tabular}{ccc}
    \hline
    \textbf{Model} &  \textbf{10\%BLiMP} $\uparrow$ & \textbf{SG} $\uparrow$ \\
    \hline
    pretrained GPT2 & 82.8 & 79.4 \\
    Post-GPT2 & 83.1 & 84.6 \\
    GiLT-GPT2 & \textbf{83.2} & \textbf{85.5} \\
    \hline
  \end{tabular}
  \caption{Results of pretrained GPT2, Post-GPT2 and GiLT-GPT2 on BLiMP and the SG test suites.}
  \label{tab:finetune sg result}
\end{table}

\paragraph{Result.} In Table~\ref{tab:finetune result}, we report F1 scores for SST2 and RTE, accuracy/F1 for MRPC, and Pearson/Spearman correlation for STS-B.  
\revisionapril{It can be seen that GiLT-GPT2 wins all tasks against Post-GPT2}, implying generally enhanced language understanding capabilities.
It is also worth noting that our parsing process uses a modest beam size of 20 (compared to 300 in language modeling), yet achieves good task performance. \revisionapril{Furthermore, GiLT-GPT2 maintains performance surpassing Post-GPT2 on both BLiMP and SG as shown in Table~\ref{tab:finetune sg result}, which reflects the strong syntactic generalization ability of GiLT.}

\subsection{Ablation Study}
\label{sec:ablation}
We design five controlled ablations: (1) --degree: removal of degree from the feature tape, (2) --depth: removal of depth, (3) --distance: removal of distance, (4) \revisionapril{--weights of degree}: removing the degree weighting coefficients $m_{in}$ and $m_{out}$, and (5) --weights of distance: removing the distance weighting coefficients $m_{in}$ and $m_{out}$. \revisionapril{Each ablation is re-trained from scratch separately}. We use GiLT-PSD as the base model and evaluate on both PPL, SG and 10\%BLiMP as in previous sections.

As shown in Table~\ref{tab:ablation}, the PPLs of the six settings are at the same level, but their syntactic generalization performances can be quite different. Notably, the --degree, --depth and --weights of distance model exhibit better 10\%BLiMP than the GiLT-PSD base model, but they yield significantly lower SG scores. The other two ablation settings degrade on both 10\%BLiMP and SG. The ablation study indicates that each feature captures a distinct linguistic aspect and their combination leads to more robust overall performance.

\begin{table}[t]
    \centering
    \small
    \begin{tabular}{cccc}
        \hline
         \textbf{Model} & \textbf{PPL} $\downarrow$ & \textbf{10\%BLiMP} $\uparrow$ & \textbf{SG} $\uparrow$ \\
         \hline
         GiLT-PSD & 14.9 & 75.7 & \textbf{79.7} \\
         --degree & 14.9 & \textbf{76.5} & 75.3 \\
         --depth & 14.9 & 76.4 & 73.8 \\
         --distance & 14.9 & 74.5 & 75.1 \\
         --weights of degree & 14.9 & 74.8 & 77.7 \\
         --weights of distance & 14.9 & 76.2 & 77.0 \\
         \hline
    \end{tabular}
    \caption{The results of the ablation study.}
    \label{tab:ablation}
\end{table}

\begin{table}[t]
    \centering
    \small
    \setlength{\tabcolsep}{3pt}
    \begin{tabular}{cccc}
        \hline
         \textbf{Model} & \textbf{Beam Size} & \textbf{Tokens/s} $\uparrow$ & \textbf{Max CUDA Memory} $\downarrow$ \\
         \hline
         TXL & 1 & 52.6 & 2.8 \\
         DTG & 1 & 24.4 & 2.9 \\
         GiLT & 1 & 43.0 & 3.6 \\
         \hline
         DTG & 20 & 5.4 & 6.4 \\
         GiLT & 20 & 24.7 & 4.2 \\
          \hline
         DTG & 100 & 0.9 & 22.7 \\
         GiLT & 100 & 9.0 & 6.7 \\
          \hline
         DTG & 300 & / & >48.0 \\
         GiLT & 300 & 3.5 & 12.9 \\
         \hline
    \end{tabular}
    \caption{The results of the generation speed test. \revisionapril{CUDA memory consumption is measured in GB.}}
    \label{tab:generation speed}
\end{table}

\subsection{Efficiency Comparison}
We have observed the strong performance of DTG in Table~\ref{tab:language modeling result}. However, DTG introduces additional structural tokens, which slows down inference. We measure the speed of DTG and our model on a small set of sentences when using beam search of dependency graphs with the same beam size $b$. We also measure the greedy decoding speed of TXL as a baseline, which does not require beam search of sentence structures and can be seen as using a beam size of 1. It can be seen that GiLT is slightly slower than TXL when $b=1$, showing that the low consumption of our extra module. Compares with DTG, our GiLT remains significantly faster and more memory efficient. As $b$ increases, the efficiency degrades and GPU memory grows, yet both worsen markedly more slowly for GiLT than DTG. Note that when $b=300$, we were unable to complete DTG’s inference on our NVIDIA A6000 GPU due to excessive memory demands.

\begin{figure*}[t]
    \centering
    \includegraphics[width=0.65\textwidth]{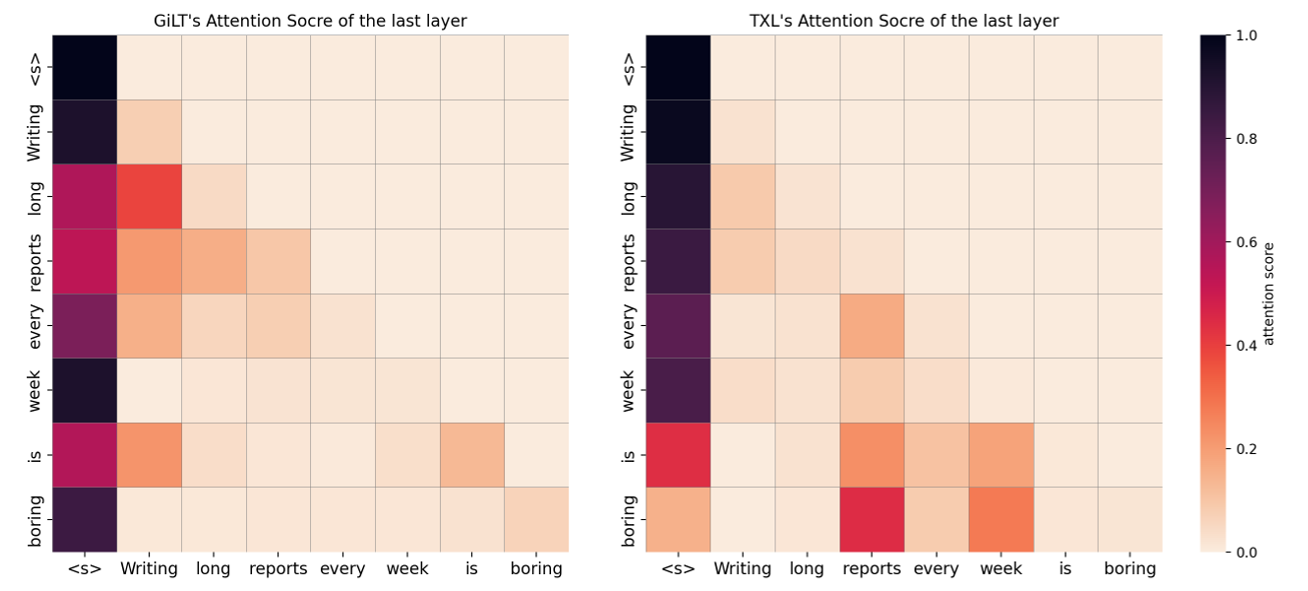}
    \hfill
    \includegraphics[width=0.27\textwidth]{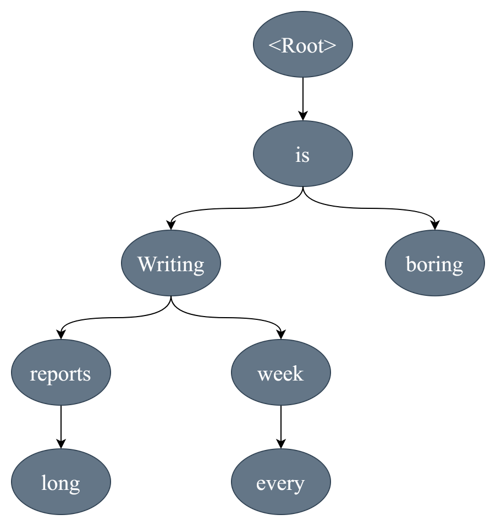}
    \caption{Left: visualization of attention scores of the first head in the last layer of GiLT (left) and TXL (right) given the input ``\texttt{Writing long reports every week is boring.}''. \revisionapril{Right: the predicted PSD dependency graph by GiLT-PSD, which also serves as the silver dependency graph of the given input.}}
    \label{fig:showcase}
\end{figure*}

\subsection{Case Study}
\label{case study}
\revisionapril{We obtain the attention scores of both GiLT-PSD and TXL and the predicted PSD graph by GiLT-PSD when inputting ``\texttt{Writing long reports every week is boring}'', and visualize them in Figure~\ref{fig:showcase}. Above all, GiLT-PSD correctly predicts every dependency of the PSD graph.} For the attention scores, TXL can be seen to consistently assign a large proportion of attention to the most recent noun and fails to identify the subject of this sentence. In contrast, GiLT correctly focuses on ``\texttt{Writing}'' when the input is ``\texttt{is}'' since ``\texttt{Writing}'' is the governing word of the subject. When inputting ``\texttt{week}'', GiLT assigns attention scores more evenly to ``\texttt{long reports}'' and ``\texttt{every week}'' besides the attention sink, since both phrases modify ``\texttt{Writing}''.

\section{Related Work}
There have been studies about leveraging recursive linguistic structural (symbolic) information for sequential language modeling. For syntactic LMs with neural architectures, RNNGs \citep{dyer-etal-2016-recurrent}, jointly model the syntactic structure and words by integrating top-down transition-based constituency parsing into a recursive neural network, while recent studies \citep{qian-etal-2021-structural, yoshida-oseki-2022-composition, transformer-grammars} have applied this approach to Transformers, which explicitly model a syntactic tree along with words by imposing hard constraints over attention masks to simulate the shift/compose operations in transition-based parsing.
\citet{hu-etal-2024-generative} further explores an unsupervised training framework for constituency-based syntactic LMs, showing the potential of training syntactic LMs at scale.

In addition to constituency-based models mentioned above, studies on those based on dependency tree structures \citep{buys-blunsom-2015-generative, mirowski-vlachos-2015-dependency}, 
also achieve improved syntactic generalization performance. A recent example is Dependency Transformer Grammars \citep{zhao-etal-2024-dependency}, which employs a constrained attention pattern similar to \citet{transformer-grammars} to encourage head-dependent representation learning.

Both constituency-based and dependency-based studies incorporate the inductive bias of symbolic structures into the self-attention mechanism by regulating the attention masks dynamically. Some other studies also focus on adapting the self-attention modules, or both \citep{wang-etal-2019-tree, peng-etal-2019-palm, deshpande-narasimhan-2020-guiding, murty-etal-2023-pushdown}, whereas our work follows the conventions of adaptation, modifying the self-attention module by incorporating dependency graph feature representations without changing the input or output space of Transformer LMs.

These models show considerable performance in generalizing syntactic information via recursion as tree structures. However, most of these studies focus solely on trees rather than a more general and flexible form: graphs. 
One notable work \citep{prange-etal-2022-linguistic} proposes a model that exploits information from both syntactic and semantic graphs. However, it only introduces graph-informed language modeling without actually modeling the explicit symbolic structure: gold syntax and semantics are needed for both training and test-time inference of the model. Semantic graphs are also employed to guide the model in other fields such as machine translation and visual tasks, but these studies directly apply the gold signals for model augmentation from semantic graphs instead of encoding the graphs into the model \citep{aue-etal-2004-statistical, ijcai2024p0472}.
GiLT differs from these models as we model graphs in the Transformer LM, and we can incrementally build a graph along with the next token prediction without graph supervision during inference.

\section{Conclusion}
We propose GiLT, a novel type of syntactic language models that incorporates dependency graphs---a more general and flexible form of linguistic structural information compared with traditional syntactic tree structures---into Transformers. GiLT jointly predicts tokens and dependencies, incrementally constructing a dependency graph and using features extracted from it to modulate attention scores. Experiments show that GiLT achieves enhanced syntactic generalization without introducing extra tokens and with minimal impact on perplexity. \revision{Additionally, finetuning GiLT from pretrained language models also improves language understanding performance on several downstream tasks.}
These results demonstrate GiLT can effectively construct dependency graphs of generated sentences and and extract their structural information to serve as inductive bias for language modeling. Our future work is discussed in Appendix~\ref{future}.

\section*{Limitations}
During inference, we rely on beam search of dependency graphs to estimate the marginalized probability, which can only provide its lower bound. Although our dependency population space is constant and independent of the sequence length, beam search of dependency graphs remains computationally expensive.

Additionally, the discussion in Appendix~\ref{sec:appendix result} suggests that the performance limitations observed on GiLT-DP are primarily due to the under-utilization of tree properties in our graph-based modeling approach. This insight highlights the potential for further research to focus on better integrating the inherent properties of graphs, such as the presence of multiple heads, to improve the model's overall performance and effectiveness.

\section{Acknowledgments}

This work was supported by the robotic AI-Scientist platform of Chinese Academy of Science, the HPC platform of ShanghaiTech University, and the Core Facility Platform of Computer Science and Communication, SIST, ShanghaiTech University.

\bibliography{custom}

\appendix

\section{Other Experimental Details}
\label{sec:appendix finetune}
\paragraph{Hyperparamter for finetuning}
To obtain Post-GPT2, we use a batch size of 64, 5000 warmup steps, a cosine decay schedule, and a maximum learning rate of 3e-5 to finetune pretrained GPT2 on BLLIP-{\small LG}. To obtain GiLT-GPT2, we assign a larger maximum learning rate of 1.5e-4 for the new parameters. Since the newly initialized parameters disrupt the semantics of hidden states, we left the language model train alone for 5000 steps before training jointly with the biaffine module using the same configuration as above. 

For downstream tasks, we use a batch
size of 64 and a fixed learning rate of 7.5e-6.
We choose the best model based on performance on the validation set. We use the following prompts to convert text classification task into language modeling:
\begin{itemize}
    \item \textbf{RTE}: We utilize the following prompt: \\
    \emph{Sentence1:\{$s_1$\}; Sentence2:\{$s_2$\}; Label:\{$l$\}}. $l \in$ \{0, 1\} for input sentence pair $(s_1, s_2)$
    \item \textbf{MRPC}: Given input sentence pair $(s_1, s_2)$, we construct the prompt: \\ \emph{Sentence1:\{$s_1$\};} \emph{Sentence2:\{$s_2$\};} \emph{Label:\{$l$\}}. $l \in$ \{\text{inequivalent}, \text{equivalent}\}.
    \item \textbf{SST2}: Given string $s$ and label $l$, prompt is: \\ \emph{Sentence1:\{$s_1$\}; Sentiment:\{$l$\}}. $l \in$ \{0, 1\}.
    \item \textbf{STS-B}: Given the sentence pair $(s_1, s_2)$, we create the prompt \emph{Sentence1:\{$s_1$\}; Sentence2:\{$s_2$\}; Score:}. We use the final hidden states to train a linear regression model, training jointly with LM.
\end{itemize}

\paragraph{Computational costs}  We use PyTorch version 2.7.0 for all experiments. For language modeling experiments, we spent one NVIDIA A6000 GPU for each training, which lasted about 50 hours. For finetuning experiments, we spent one NVIDIA H800 GPU for each training, which lasted less than 1 hour for each task.

\section{Discussion on different parsing}
\label{sec:appendix result}
By analyzing metrics in Table~\ref{tab:language modeling result}, we discover that the order of performance in perplexity of models trained on different datasets can be listed high to low as: PSD, DM, PAS and DP. It also roughly conforms to this order in other metrics. 

\begin{table}[h]
    \centering
    \small
    \begin{tabular}{ccccc}
         \hline
         \textbf{SDP Dataset} & \textbf{Avg. Dependencies} & \textbf{PPL} \\
         \hline
         PSD & 16.9 & 14.9 \\
         DM  & 18.8 & 14.9 \\
         PAS & 24.6 & 15.0 \\
         DP  & 24.2 & 15.5 \\
         \hline
    \end{tabular}
    \caption{Average number of dependencies per sentence in different SDP dataset based on BLLIP-{\small LG} and reported perplexity of each model from Section~\ref{exp:lm}.}
    \label{tab:count}
\end{table}

We calculate the average number of dependencies in the graphs and report the results in Table~\ref{tab:count}. We can surprisingly find that the fewer dependencies we need to establish, the better performance we will get. This is likely because fewer dependencies result in less noise we obtained from silver dependency graphs, and the simpler graphs are probably easier to model. 

The exception is GiLT-PAS with better performance than GiLT-DP when PAS has average dependencies more than DP.
Performance degradation on the DP dataset is not unexpected, as the dependency graphs for DP are essentially trees: we lessen the constraints in the models, hence our model for dependency trees are weaker to leverage the unique recursive properties of trees. This suggests that while GiLT is able to handle more dependencies in graphs with relatively minor performance degradation, it has limitations in effectively utilizing tree structures, a specific type of graph.

\section{Future work}
\label{future}
For future work, we plan to explore the potential of the feature tape for jointly modeling multiple types of dependency graphs. This presents a significant challenge for both effective training and efficient inference. Furthermore, we consider unsupervised training for GiLT as another promising direction.
\end{document}